\documentclass[runningheads]{llncs}
\usepackage[utf8]{inputenc}
\usepackage{graphicx}
\usepackage{enumitem}
\usepackage{amssymb}

\usepackage{hyperref}
\hypersetup{colorlinks=true}

\newcommand{\indep}{\perp \!\!\! \perp}

\begin{document}

\title{Fuzzy Stochastic Timed Petri Nets for Causal properties representation}
\author{Alejandro Sobrino\inst{1}\and
Eduardo C. Garrido-Merch\'{a}n\inst{2}\and
Cristina Puente\inst{3}}
\date{September 2020}

\institute{Universidad Santiago de Compostela, Galicia, Spain
\email{alejandro.sobrino@usc.es} \and
Universidad Aut\'onoma de Madrid, Madrid, Spain
\email{eduardo.garrido@uam.es} \and
Universidad Pontificia de Comillas, Madrid, Spain
\email{cristina.puente@icai.comillas.edu}}

\maketitle

\begin{abstract}
Imagery is frequently used to model, represent and communicate knowledge. In particular, graphs are one of the most powerful tools, being able to represent relations between objects. Causal relations are frequently represented by directed graphs, with nodes denoting causes and links denoting causal influence. A causal graph is a skeletal picture, showing causal associations and impact between entities. Common methods used for graphically representing causal scenarios are neurons, truth tables, causal Bayesian networks, cognitive maps and Petri Nets. Causality is often defined in terms of precedence (the cause precedes the effect), concurrency (often, an effect is provoked simultaneously by two or more causes), circularity (a cause provokes the effect and the effect reinforces the cause) and imprecision (the presence of the cause favors the effect, but not necessarily causes it). We will show that, even though the traditional graphical models are able to represent separately some of the properties aforementioned, they fail trying to illustrate indistinctly all of them. To approach that gap, we will introduce Fuzzy Stochastic Timed Petri Nets as a graphical tool able to represent time, co-occurrence, looping and imprecision in causal flow.
\end{abstract}
\section{Imagery and causality}
Since ancient times, imagery has been used to illustrate concepts and actions \cite{tye2000imagery}. In the very beginning, even if in a rudimentary form, cave paintings schematically represented cognitions about hunting or agricultural
activities \cite{bahn1998cambridge}. In latest times, automata theory or Chomsky syntagmatic grammar are examples of imagery
tools dealing with abstract theories and deep concepts \cite{cook2014chomsky}. Thus, the Turing machine as a central computing
unit moving left and right on an infinite tape according to a program pictorially illustrates the universal concept of computability \cite{shannon1956universal}. In the same way, Chomsky’s syntagmatic trees illustrate the geometry of the sentences, depicting the node dominances that a flat description is not able to portray.

Imagery language is often opposed to verbal sentences, although both aim to mirror the structure of our
thoughts \cite{paivio1978imagery}. Oral language, and especially its complement writing, was linked to the social demands of
exchange control, codifying laws or recording history in a reliable and precise way. In the scientific area,
the relevance of writing was reflected in the name adopted by a prominent school in Philosophy of
Science, e. g., ‘the statement view’. But as Popper said (1963:28) \cite{popper2014conjectures}: “although clarity is valuable in itself, exactness or precision is not”. Under the influence of iconoclastic religions, words have been considered
since the Renaissance as the precise way to express thought, but lately the role of images as a source of
clarity has been vindicated. Thus, for the Larkin \& Simon view \cite{larkin1987diagram}, imagery has
advantages, favoring gathering information about a single element by perceiving at a glance its role in the
net, and so, facilitating perceptual inferences, quite familiar to human beings and generally difficult to
obtain by other means. Imagery and written language work in different ways: while imagery illustrates
topological relations, texts show sentence concatenations, but both contribute to the knowledge
acquisition, as Paivio \cite{paivio2006mind} pointed out in his Dual Coding Theory. In his view, cognition uses two different subsystems: the imagery one, based on the observation of objects and events and the
relations among them, and the verbal one, attaching the representations in texts.

Aristotle advocated imagery as a tool for representing any kind of knowledge, being concrete or abstract \cite{beare2010memory}. On the contrary, Kant supported that images are suitable to represent perceptual information, but not
abstract ideas \cite{smith2011immanuel}. Between them, a new point of view can be considered: the third-person imagery \cite{nigro1983point}. That approach changed the focus from a concrete or individual view to a general or collective one, demanding a look from an third or neutral observer, and so validating the role of
imagery supporting abstract thoughts. We join the Aristotle view about the functional role of imagery
representing human knowledge, being concrete or abstract if the third-person view is adopted. In this view, images have a double role: they outline the essential traits of knowledge and make easier to remember them \cite{paivio2006mind}.

Causality, a main characteristic of science, makes an extensive use of images through graphs, a kind of
pictures showing the cause-effect link typical of many human cognitions or explanations \cite{fiske1979imaging}. Agreeing with the third-person imagery, causal graphs permit to represent type and token causal relations, the latter considered as prototypes of the particular problems they illustrate. Causal graphs, a
kind of imagery representation, provide insight views of causal problems, separating the wheat from the chaff, and supplying adequate explanations of intricate puzzles, illustrating how a node is causally reached from the parent ones.

Causality is about causal relations, frequently addressed mathematics, philosophy, ecology, economy or psychology \cite{berzuini2012causality}. These subjects used verbal descriptions or mathematical
equations to gather causal knowledge. But graphics are not an exception: Philosophy or computational
approaches make an extensive use of them. In the sequel, we will show the role of graphs explaining
some causal properties associated with some well-known causal puzzles.

\section{Properties of causal relations: time, concurrency, feedback and imprecision.}

A causal relation, $e = r(c) : c \in U_c, e \in U_e, U_c \to U_e$, represents a link between the cause $c$ (simple or complex that belongs to some set $U_c$ of causes), and the effect $e$, provoked by the
cause that belongs to a set $U_e$. The intersection between sets $U_c$ and $U_e$ can be void $\emptyset$, can have some elements in common or both sets can be equal $U_c = U_e$, depending on the problem. Given the representation of a particular problem, we could have that an effect $e_j$ of a cause $c_j$, modelled by the causal relation $j$, can be the cause $c_i$ of another effect $e_i$ of a causal relation $i$. Hence, causes $\mathbf{c}$ and effects $\mathbf{e}$ are properties of objects $\mathbf{o}$. From now on, we will talk about objects, where being a cause or effect is a semantic property of them. Causality is often represented by causal graphs. A causal graph is a pair $G=(V,E)$ where $V$ is a set whose elements are called vertices, that are the previously defined objects, and $E$ is a set of causal directed relations between pairs of objects that represent the causal semantic information of these objects. $E$ is, for a given problem, then a subset of the relations given by the set of all possible relations between all the objects in a problem $U_e = r(U_c)$. A graph is a visual tool for illustrating associations
between objects using nodes and directed arrows. The arrows show the link between the cause and the
effect, and its direction, the flux of the causal influence. Graphs illustrate paradigmatically the connection from cause $U_c$ to effect $U_e$.

If we define a causal relation by an edge $e$ of the causal graph, this object has four different properties. Inter alia, causal relations $E$, representing real world problems, are sensitive to time $e_t$, concurrency $e_c$, feedback $e_f$, and imprecision $e_i$. We define these properties below.

\subsection{Time}
Let us consider that $X$ is a cause of $Y$ given a relation $r$. A causal relation $r(X,Y)$ is also a function of time, $T$. That is: $r(X,Y,T)$, where $T$ is the time spent between the event $X$ generates the event $Y$. Regarding time, according to Taylor \cite{taylor1992formal}, causal relations ,$E$, satisfy the following properties: Irreflexivity: \textit{Nihil is causa sui.} That is, we can not have $X \to X$, or $r(X,X,T)$, because the sequence would be infinite. Anti-symmetry: Provided that $X$ and $Y$ are distinct, if $r(X,Y,T)$, then we can not have $r(Y,X,T)$, again, because the sequence will also be infinite. Precedence: If $r(X,Y)$, then $X$ precedes $Y$. That is, $X \in pa(Y)$.

Precedence in causal relations shows the arrow of time: from cause to effect and not the reverse. Even if time is a key factor in causation, it is important to remark that it is not always considered by the usual graphical tools that represent causality, as we will see later.

\subsection{Concurrency}

Frequently, an effect $Y$ demands the simultaneous presence of several causes; i. e., causality is often pluri-causality and concurrency is a type of that. We will denote the set of direct causes $\mathbf{X}$ of an effect $Y$ as the one retrieved by the function $pa(Y)$. We define that a cause $X$ is direct of an effect $B$ if a single edge is needed to link both objects in a causal graph $G(V,E)$. In this paper, we use concurrency to denote co-occurrence of causes.
Plural causality was paradigmatically addressed by INUS causality \cite{mackie1974cement}. Typically, an effect $Y$ has a lot of causes $\mathbf{X}$, some of them possibly grouped into clusters $\mathbf{X}_j$. Each cluster $\mathbf{X}_j$ is sufficient, but not
necessary, to provoke the effect $Y$, that is, $r_j(\mathbf{X}_j,Y) \forall j$ where $j$ is the index of a cluster. Causes $\mathbf{X}_j$ are groups of single factors $X$ and each factor $X$ is an INUS (Insufficient but Nonredundant part of an Unnecessary but Sufficient) condition for the effect $Y$. Each
cluster $\mathbf{X}_j$ is a set of concurrent factors $X$ able to provoke the effect $Y$. At the end we have that pluri-causality relations $r$ can be all represented as: $\lor_{j=1}^{J}(\wedge_{i=1}^I X_i)_j\to Y$ where $X_i$ is an INUS, $(\wedge_{i=1}^{I_j} X_i)_j = \mathbf{X}_j$ is a sufficient, but not
necessary cluster, $J$ is the number of clusters and $I_j$ is the number of INUS for a cluster $j$. Concurrency of causes
is typical in causal scenarios, but it is not addressed by all causal graphs, as we will show below.

\subsection{Feedback}

A causal process $r(\cdot, \cdot)$ frequently shows mutual influence. If an element $X$ influences itself directly or indirectly, this is referred to as a causal loop, i.e., a closed cause/effect feedback $X \to \cdot\cdot\cdot \to X$. There are two main classes of causal loops: In positive feedback loops, the loop leads to a reinforcing feedback as cycles increase more and more. Consider $f(x),g(y)$ as functions of random variables and the graphical model $f(x) \to g(y) \to \cdot \cdot \cdot \to f(x)$, where $\cdot\cdot\cdot$ can be from zero to an indeterminate number of functions. If we define $i \in \mathbb{Z}$ as the $i$ iteration of the execution of every function of the graphical model $f$ and $g$, then, the positive feedback loop will always comply, for every one of its functions $f$: $f(x|i=j+1) > f(x|i=j)$. On the other hand, in negative feedback loops, the loop $f(x) \to g(y) \to \cdot \cdot \cdot \to f(x)$ seeks an equilibrium balancing the fluctuations in the output: if the output increases, the causal loop pushes the input value down and if it decreases, it pushes the value up. \textit{Id est}, $\mathbb{E}(f(x|i=j+n)) = f(x|i=j)$ where $n,j \in \mathbb{Z}$. 

A special type of negative feedback is the one that involves a time delay. A delay happens when it takes time before the effect plays out. In this case, the feedback signal can arrive later, turning a positive feedback into a negative one. E. g., sales increase orders, and more orders favor more sales, but if orders are delayed (dotted line), sales may fall. Time delay breaks the balance. Feedback is prevalent in economic or ecologic scenarios, where the notions of reinforcement or balance have a relevant role.
Finally, another property emerges if causation is empirically scrutinized:

\subsection{Imprecision}

Causes $\mathbf{X}$, effects $\mathbf{Y}$ and cause-effect links $r(x,y) : x \in \mathbf{X}, y \in \mathbf{Y}$ are often qualified by degrees $\epsilon \in [0,1]$ (percentages) or if the causal relation representation is a text: adverbs $\mathbf{w}$. These adverbs can be modelled by point estimations $\epsilon \in [0,1]$ or, as we will later describe, by probability distributions $p(w)$ or fuzzy functions $f(w)$. These objects denote
a measure of frequency, intensity, or strength, that can be represented by point estimations or different probability mathematical objects. Let us define a causal sentence as a text representation $T$ that includes a semantic meaning involving a causal relation between objects $r(X,Y)$. Causal sentences $\mathbf{t}$ automatically mined from the short papers included in Hawking’s Physics Colloquium
\cite{puente2010extraction} show that, even in hard science, causal descriptions are not crisp, but imprecise, in so far as the causal links are frequently tinged with semantic hedges or vague adjectives, as ‘too’, ‘definitely’ or ‘nearly’. Thus, physical causality is frequently rather approximate than categorical. As we will further see, we can model these adverbs $\mathbf{a}$, by point estimations $\epsilon \in [0,1]$ representing an adverb by a precise degree of uncertainty. As language is imprecise and depends on context, the time frequency adverb $a$, given by its signficant, may provide a plethora of significances, that is why Garrido \cite{garridomerchn2020uncertainty} represents these adverbs by probability distributions $\mathbf{p}(\mathbf{a}|\Theta)$, as we will further see.

Despite the empirical support, vagueness or imprecision have not always accommodation in the typical graphs representing causality. Concurrency, time, looping and imprecision seem to be usual and relevant aspects involved in many causal processes but insufficiently addressed by graphical methods. Neurons, bayesian networks, cognitive maps or probabilistic causal graphs supply tools for depicting positive, negative, proximate, uncertain, distant and single or plural causality, but neurons exclude imprecision, bayesian networks do not take time into account, cognitive maps do not consider independence in concurrent causality and probabilistic causal graphs do not consider feedback loops.
In this paper, we will propose the
Fuzzy Stochastic Timed Petri nets (FSTPN) \cite{liu2016fuzzy} as a tool that can contribute to represent
co-occurrence, time, feedback and imprecision in causal flux. Although co-occurrence is an aspect
inherent to basic Petri Nets (PN) (Murata, 1989), time was only incorporated in a later evolution of them
known as Timed PN (TPN) \cite{molloy1985discrete}. It should be mentioned that, although TPN still lack
mechanisms to deal with imprecision, FSTPN can represent the vagueness involved in the probability of
the transitions firing. Thus, FSTPN seem to be a useful tool for representing concurrent systems
accounting for time parameters and fuzzy probabilistic causal influences.
To accomplish this aim, the rest of the paper is structured as follows: First, traditional visual methods for representing causality are presented. Then, we illustrate deficiencies of the aforementioned methods. We provide a fuzzy stochastic timed Petri nets overcome these issues. Lastly, we summarize the contributions of the proposed method.

\section{Graphical methods for representing causal relations.}

We introduce a review of graphical causal methods that represent some of the properties of causal relations that we have described previously.

\subsection{Neurons.}

Neurons were firstly considered by Lewis \cite{lewis1986postscripts} as a way to represent causal relations $c(\cdot, \cdot)$. A neuron $N$ is an oriented graph $N(V,E)$ with nodes $V$ denoted by circles symbolizing the cause $x \in V$ or the effect $y \in V$ and causal links $\cdot \to \cdot$ depicting the causal influence $X \to Y \forall X,Y \in V$. Circles are usually labeled with lower case letters. Circles may be shaded or not. If shaded, that we will represent as a normal cause $X$ they are actives and, thus, ready to transmit a causal influence $X \to Y \forall X,Y \in V$. If not (in blank, that we will represent as $\overline{X}$) they are inoperative and cannot pass any influence to the connected nodes $V$. Links between nodes $\cdot \to \cdot$ are represented by arcs $\to$, which can end in an arrow, which we will represent as $\cdot \twoheadrightarrow \cdot$, or in a circle, that we will represent as $\cdot \multimap \cdot$. If ended in an arrow, a connection $X \twoheadrightarrow Y$ is stimulatory; if finished in a circle $X \multimap Y$, inhibitory. Let us suppose an arrow that connects a node $X$ with a node $Y$, that is $X \to Y$. The arc
stimulates $X \twoheadrightarrow Y$ or inhibits $X \multimap Y$ only if $X$ is active, not inactive $\overline{X}$. If $Y$ receives both a stimulatory $X \twoheadrightarrow Y$ and an inhibitory arrow $Z \multimap Y$, inhibition cancels stimulation. Briefly, neurons satisfy the following principles: A neuron $X$ is activated if it is stimulated by at least another neuron $Y \twoheadrightarrow X$ and is not inhibited by any other neuron $Z \multimap Y$. A neuron $X$ is not activated $\overline{X}$ if it is inhibited by one or more neurons $\mathbf{Y} \multimap X$. Lastly, the start neurons have no outside connections. They are active if they are shaded.

Neurons admit neither time nor imprecision and although they accept pluricausality, using the syntax of neurons we cannot force the concurrence of two or more nodes causing the effect. If a node is active, it can cause the effect by itself, the other node is not
necessary. But in many causal scenarios, the joint contribution of two or more causes for reaching the effect is demanded.

\subsection{Boolean functions.}

Representing cause $X$ and effect $Y$, Boolean functions $f(\mathbf{b})$ involve binary values $\mathbf{b}$ for variables and logic formulas for modeling indicative or factual causation. Variables are supposed to be independent of each other, $X \indep Y$, and the causal influence is calculated in terms of the present (1), $X$, or absent (0), $\overline{X}$, causes, provoking (1), $Y$ or not (0)
the effect $\overline{Y}$. For example, let us suppose that a surgical team consists of 4 persons $(A, B, C, D)$, one of them being the chief surgeon ($A$). The decision to perform a surgery on a patient ($S$) is made by simple majority and the vote of the chief surgeon has a double value in case of a tie. The following formula f($\mathbf{b}$) represents the approval of the action:

($\neg$A$\wedge$B$\wedge$C$\wedge$D)$\lor$(A$\wedge\neg$B$\wedge\neg$C$\wedge$D)$\lor$(A$\wedge\neg$B$\wedge$C$\wedge\neg$D)$\lor$(A$\wedge\neg$B$\wedge$C$\wedge$D)$\lor$\\
(A$\wedge$B$\wedge\neg$C$\wedge\neg$D)$\lor$(A$\wedge$B$\wedge\neg$C$\wedge\neg$D)$\lor$(A$\wedge$B$\wedge$C$\wedge\neg$D)$\lor$(A$\wedge$B$\wedge$C$\wedge$D).

So, the action is caused if $A$ does not vote positively but $B$, $C$ and $D$ do or if $A$ and $D$ vote positively but $B$ and $C$ do not. The previous canonic formula is a maxiterm that can be shortened into an equivalent one using minimization algorithms, as the Quine-McCluskey method \cite{ragin2014comparative}
or the Karnaugh maps \cite{karnaugh1953map}. Boolean functions $f(\mathbf{b})$ do not show dependencies between indirect causes nor approach time or imprecision.

\subsection{Causal Bayesian networks.}
Bayesian networks are probabilistic graphical models $G(V,E)$ that use Bayesian inference for the computation of the probability of an effect $Y$ given causes $\mathbf{X}$ where all the causes and effects are modelled by random variables \cite{jensen2007bayesian}. Concerning the described properties of causal relations introduced in the previous section, Bayesian networks try to overcome the limitations of Boolean functions mentioned above by: Detecting the indirect influence of a cause $X$ in the rest of the graph $G$. Representing imprecision or uncertainty, using random variables $p(x)$, generally associated to Gaussian distributions. Checking independence between variables in the causal process. Concretely, the absence of edges $E$ on bayesian networks model conditional independences between causes and effects. 

Indirect influence depends on the dependency diagnoses between random variables, changing according to: The interventions made on the graph $G$ and the structure of the graph given by its edges. An edge between a cause $X$ and an effect $Y$ is modelled by a factor $P(Y|X)$ of the joint distribution modelled by the Bayesian network.

A Bayesian network represents independence of events using conditional probabilities $P(\cdot|\cdot)$. Formally, a Bayesian network is a directed acyclic graph (DAG) $G(V,E)$ that models a joint distribution of conditional distributions. The nodes $V$ represent events that are modelled by random variables and the directed arcs $E$ represent causal relations or factors that appear in the joint probability distribution. Nodes are labeled with variables (capital letters) and variables can be instantiated to data, by sampling over the probability distributions $p(\cdot)$ that are associated to the random variables modelled by the nodes. The root nodes (parent nodes $pa(X)$ of a random variable $X$) are modelled with \textit{a priori} probabilities $p(X|\theta)$, where $\theta$ is the set of \textit{a priori} parameters of the probability distribution $p(X)$ and the children nodes with conditioned probabilities $p(Y|pa(Y))$, which represents the probability
of a node conditioned upon its parents $pa(\cdot)$. Independence is adequately illustrated by conditional
probability: $X$ is probabilistically independent of $Y$ conditioned on $Z$ if $\forall X, Y, Z, P(X|Y, Z)=P(X|Z)$; e. g., once
the $Z$ value is known, the value of $Y$ does not change the probability of $X$. But independency diagnosis can vary, \textit{ibid}, if interventions are performed on the nodes, as we will further observe. 

Consider a graph representing the following causal relations $B \to A$, $C \to A$, $A \to E$. The effect $A$ is represented by the conditional probability $p(A|B,C)$ , \textit{id est}, $A$ is caused by $B$ or $C$, having $B \indep C$, since there is no arrow from B to C or \textit{vice versa}. Nonetheless, we can conjecture that there be some relationship between $B$ and $C$ because both events are causes of $A$. Suppose that we check $A$ and observe that it happens. In that case, knowing the value of one of the causes (e.g., $B$) informs us about the other cause
($C$), opening now a communication between them. 
Even though causal Bayesian networks deal with imprecision through probability distributions and independency diagnoses in causality, they do not address time and feedback influence. But causal loops are usual in social and ecological
scenarios, where, in addition, time plays a key role.

\subsection{Fuzzy cognitive maps.}

First approached by Tolman \cite{tolman1948cognitive}, a cognitive map $M$ is a graphical tool for the spatial representation of a
situation, favoring, at a quick glance, the identification of the items $\mathbf{i}$ involved in the scene as well as the
links $\mathbf{e}_{ij}$ between them, where $e_{ij}$ is a link from the item $i$ to the item $j$. Later, Axelrod \cite{axelrod1976cognitive} used cognitive maps to represent political scenarios with
various events or agents causally linked. Kosko \cite{kosko1986fuzzy} extended cognitive maps to fuzzy cognitive maps in order to host imprecision and uncertainty, characteristic of many daily settings. In so far as the
multiple-valued or fuzzy cognitive maps include the Boolean ones as a particular case, next we will
address the more general case.
Fuzzy cognitive maps are graphs $G(V,E)$ with nodes $V$ labeled by concepts. A concept $v_i$ consists of a modifier, often
an adjective or a noun, and a quantity (or its negation), usually an adverb. For instance, social {stability,
instability} can cause {increasing, decreasing} of prices, being {stability, instability; increasing,
decreasing} quantities and {social, prices} modifiers. Like Bayesian networks, fuzzy cognitive maps
illustrate causal influences $r(v_i, v_j, c)$ from a cause $v_i$ to an effect $v_j$ decreasing, being neutral or increasing, which is given by the categorical variable $c = [+,-,\cdot]$, but unlike Bayesian networks, they score the causal impact using ordinate values $[+,-,\cdot]$, and not a measure of probability $b \in \mathbb{R}^{[0,1]}$. As it has been described, influence admits three valuations: (a) positive (favoring the effect, $+$), (b) negative
(inhibiting the effect, $-$) or (c) neutral (causing no influence on the effect, no mark). Thus, initially, fuzzy cognitive maps used a trivalent logic, a special type of fuzzy logic for representing the causal influence.

Kosko \cite{kosko1986fuzzy} expanded the qualification of the causal influence from three-valued logic $[+,-,\cdot]$ to infinite valued
one and later, to a linguistic-valued fuzzy logic.
Regarding infinite valued logic, the links can take values in the interval $e \in \mathbb{R}^{[-1, 1]}$: the sub-interval $e_p \in \mathbb{R}^{(0, 1]}$
denoting positive causality, the interval $e_n \in \mathbb{R}^{(-1, 0]}$ negative causality and the value $0$ the absence of causal
influence. Thus, $-0.2$ denotes a small negative causal influence and $0.6$ means a rather positive causal
influence.

Values are aggregated using t-norms and t-conorms, a class of binary operators to model conjunction $ \cdot \wedge \in \mathbb{R}^{[0,1]} \cdot$ and
disjunction $ \cdot \lor \in \mathbb{R}^{[0,1]} \cdot$ in multiple-valued and fuzzy logic (Nguyen, 2006). Typical norms are $\min(x,y)$ (minimum or
G\"odel t-norm), $x\cdot y$ (product t-norm) or $\max(x+y-1,0)$ (Lukasiewicz t-norm) and typical t-conorms are
$\max(x,y)$ (maximum or G\"odel t-conorm), $x+y-x \cdot y$ (product t-conorm, probabilistic sum) or $\min(x+y,1)$
(Lukasiewicz t-conorm, bounded sum).

The most genuinely fuzzy extension of fuzzy cognitive maps was the linguistic-valued one \cite{kosko1986fuzzy}. Now, the links $E$ between nodes $V$ are labeled with fuzzy quantifiers $q \in R^{[0,1]}$ that are associated with an adverb $a_i \to q_i$ (many, most, a lot, etc.) that are
aggregated using t-norms and t-conorms quoted above. 

Managing linguistic values, fuzzy-valued cognitive
maps facilitate knowledge representation and inference in a human style, furnishing a flexible and
realistic tool for handling vague causal influence.
Ecological or political systems, areas of frequent fuzzy cognitive maps applications \cite{kosko1988hidden}, are largely dependent on time. For instance, geopolitical influence systems are chronologically
dependent. Fuzzy time cognitive maps, first approached by \cite{hagiwara1992extended}, included temporary
annotations $t_{xy}$ in relations between nodes $X \to_{t_{xy}} Y$, denoting a delay $t_{xy}$ before the effect $Y$ is reached.

In fuzzy time cognitive maps, directed arcs $X \to Y$ can denote positive $\mathbb{R} \in [0,1]$ or negative causal influence $\mathbb{R} \in [-1,0]$ from the cause $X$ to the effect node $Y$, and the delay is represented by $t \in \mathbb{Z}$. Numbers denoting time delays can be
understood to represent time according to units as hours, days, etc.
Fuzzy time cognitive maps illustrate imprecision, loops and time in causal links, but not concurrency,
while there is no way to ask the joint action of two or more causes for achieving an effect.

\subsection{Probabilistic causal graphs for representing causal text sentences.}

Text causal sentences are representation that often involve a set of causes $\mathbf{X}$ and effects $\mathbf{Y}$ and a time frequency adverb $a$ that expresses how probable is that, given the set of causes $\mathbf{X}$, the set of events $\mathbf{Y}$ occur. We can also represent that information, as we have seen, in a probabilistic causal graph $G(V,E)$. 

An important issue happens when we have several text causal sentences concerning the same cause and effect but a different time adverb $a$. If we want to build a knowledge graph that represents the average time frequency of the event $Y$ given by the cause $X$, that is $p(Y|X)$, and we have processed text causal sentences with different adverbs, then, how do we properly represent that uncertainty? If the same adverb $a$, depending on the context, maps to a different degree of uncertainty, how do we properly represent that? 

Garrido \cite{garridomerchn2020uncertainty} proposed an ad-hoc approach by using a slight modification on Bayesian networks to deal with an accurate model to represent that information, applying it to the detection of fake news \cite{garrido2020fake}. In a first step \cite{merchan2019generating}, a process retrieved all the causal sentences $\mathbf{w}$ of a text and stored the most representative ones as cause, effect and modifier tuples $(c,e,m)$ \cite{puente2013creating} \cite{puente2017summarizing}. With that tuples $\mathbf{w}$, we can build a weighted graph $G(V,E)$ where causal relations $r(c,e,m)$ are weighted by a quantity $m \in \mathbb{R}^[0,1]$ representing uncertainty. The model let us compute the probability of an event $Z$ that was not directly a cause of an effect $X$ but an indirect one linked by other set of causes. For example, if we have $X \to Y \to Z$, we can compute $p(Z|X,Y)$ as $p(Z|X,Y) = p(Z|Y)p(Y|X)$. More generally, $p(Z|X_n,...,X_1) = p(Z|X_n) \prod_{i=1}^{n-1} p(X_{i+1}|X_{i})$. The difference with Bayesian networks is that we are specifically modelling the probability of the connection between two causes and not the probability of the cause as if it was an event. In bayesian networks, we are modelling $p(X)$, here, we just assume that $p(X)=1$.

The previous approach was enhanced \cite{garridomerchn2020uncertainty} as the probability of a link was being modelled by a point estimation $m \in \mathbb{R}^[0,1]$ and, depending on the context where the adverb appear, this $m$ can vary. To model this uncertainty, we associated to each adverb $a$ a probability distribution $p(m)$, typically a Gaussian, associated $m$ to the mean of the distribution and also providing a standard deviation to represent the uncertainty of the adverb. Each link $p(Y|X)$ is a joint probability distribution of all the tuples $\mathbf{t}$ retrieved from text where we can find an effect $Y$, cause $X$ and an adverb $m$ that is converted into a probability distribution $p(m)$. Then, $p(Y|X) = \prod_{i=1}^{n} p(Y|X, m_i)$, \textit{id est}, we are marginalizing the adverbs, generating a joint probability distribution of all the causal sentences involving $Y$ and $X$. We can still compute $p(Z|X_n,...,X_1) = p(Z|X_n) \prod_{i=1}^{n-1} p(X_{i+1}|X_{i})$ as in the previous approach or sample a point estimation of the uncertainty of an indirect effect by sampling the first cause relation and then following the link of events. 

The new approach can represent time, in the sense that a new causal text modify each link $p(Y|X)$ by adding a new factor $p(Y|X,m_i)$ but not by having a representation $t$ of the delay. This approach does also not solve feedback lops, as it assumes that there does not exists tuples as $r(c,e,m)$ in the texts. Having reviewed all the most important causal graphical models, we have shown that none of them accurately model the properties mentioned in Section 2, being necessary the proposal of a new graphical causal model that represents them.

\section{Causal puzzles refractory to be adequately represented by standard graphical tools.}

In spite of the fact that concurrency, time, loops and imprecision are frequent in classical causal puzzles, traditional graphs show limits representing them. In this section, we mention some examples illustrating the deficiencies of the described methods. Regarding concurrency and time, neurons have the problem of symmetrical and asymmetrical overdetermination. Concerning imprecision and time, causal Bayesian networks have the fizzling and trumping issues. 

Let us describe the symmetrical overdetermination problem. A causal scenario is over-determined when the effect is caused jointly by two causes \cite{schaffer2003overdetermining}.
Suppose that, for being cured, the joint action of two drugs $a$ and $b$ are needed. For instance, for curing
Helicobacter Pylori, clavulanic acid and amoxicillin must be administered ($a\wedge b\to c$). According to the posed problem, $c$ demands the concurrency of two causes ($a$ and $b$) to be activated. But
attending the definition of neuron, it is not possible to guarantee that. Since $a$ and $b$ nodes are active, the link is stimulatory in both cases and $c$ is activated independently by $a$ or by $b$. 

On the other hand, we have the asymmetrical overdetermination issue. A situation is asymmetrical over-determined when a cause $c$ preempts the other to provoke the effect $e$ \cite{hall2003causation}. For example: Let $a$ and $b$ be two incompatible drugs, each of which is sufficient to mitigate an illness. 
Choosing one of the two drugs means forgetting the other, avoiding thus possible adverse reactions. The effect may be caused by two factors inhibiting each other; so, if they were activated at the same time, the effect vanishes. Time and anticipation are essential in this example to achieve the effect on a track. But anticipation is not considered by neurons.

Having reviewed the problems with neurons, now we illustrate issues of causal Bayesian networks. Imprecision in causal flux are addressed by Bayes Nets in a probabilistic frame. Suppes \cite{suppes1970Probabilistic} coined
positive causality posing that a cause provokes an effect if it increases its probability: \textit{i.e.}, $C$ causes $E$ if $P(E|C)>P(E)$. This postulate has been questioned and the following are some of the most common objections. They fall mainly into two categories. First, fizzling, where a cause may be less likely than other one and yet being the cause of the effect. Second, trumping, where a cause may be as likely as the other one and nevertheless failing to contribute at all to the effect.

The first case, known as fizzling, refers to a very probable causal factor that contributes to the effect
without being its cause. To illustrate this puzzle, a variant of the preemption story was presented: $a$ is a quiet and responsible person and $b$ a very known vandal. Both $a$ and $b$ have a stone in hand and a
lamppost in front of them. The probability that the lamp suffers a damage (D) is greater in the presence of b that by the presence of a. In any case, the probability that b breaks the lamp is less that 1 and the probability that a does the same is greater that 0. If a throws the stone, inhibit the action of b, who
presumably will desist from throwing his stone if the lamppost is already broken. Let suppose that, in an
unexpected rage behavior, a throws the stone against the lamppost breaking it. Although it was more
likely that b broke the lamp, it was a who did. Of course, the fizzled disposition of b it was not the cause
of the breakage of the lamppost, but its presence increased the probability of such event. Moral: the
presence of b favors the effect, but is not sufficient for its causation. Anticipation is determinant for the
causal assignment.
In the previous story, Bayesian Nets do not consider temporal precedence.

Trumping shows that even two causes being equally likely to produce the effect, the cause that happens
before surpass the other, becoming the actual cause. The effective cause is sufficient, but not necessary,
for causing the effect.
The next story illustrates trumping \cite{schaffer2001causes}: In a magical land are two wizards, Merlin (Me) and Morgana (Mo). Each of them can throw a spell on the Prince, turning him, at midnight, a frog. The laws of magic say that the first spell to occur during the day will be the one that causes the effect. And it is a matter of fact that Merlin casts his spell ($S_{Me}$) in the morning and Morgana ($S_{Mo}$) in the
afternoon. At midnight the Prince turns into a frog ($F_P$) and there is no doubt that Merlin's spell was the cause. Merlin's action did not disable the Morgana’s one; if one of them has no effect, the other does.
Although both are equally likely to cause the effect, Merlin preempts Morgana acting before. Merlin is
the cause even if he had the same probability as Morgana to provoke the effect. Using Bayesian causal
networks, a possible representation of this puzzle is shown in Figure \ref{fig:trump}, left:

\begin{figure}[htbp]
\centering{
        \includegraphics[width=3cm]{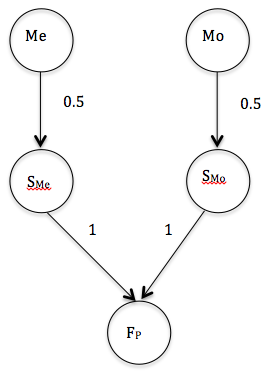}
        \includegraphics[width=3cm]{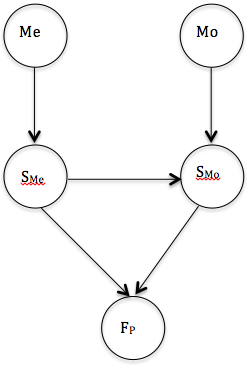}
}
\caption{Causal network representing trumping (left). Depicting causal influence from the node $S_Me$ to the node $S_Mo$ (right).}
\label{fig:trump}
\end{figure}

This network is an example of converging connection or explaining away. Recall that causal converging links say that if a node conclusion changes certainty because receives
evidence, it opens the communication between its parents. Related to the above graph, if about $F_P$ we only know that it may be caused from $S_{Me}$ or $S_{Mo}$ , their parents are independent, \textit{i.e}, to have evidences about
one of them as a possible cause do not change the certitudes about the other (knowing that Merlin casts
the spell does not indicate anything about Morgana's behavior). But if some evidence about $F_P$ is provided and information about one of the causes is available, the potential influence of the other cause can be reassessed. In the above graph, if we know that the Prince is now a frog and that is due to the Merlin’s spell, then the confidence in Morgana as a possible cause diminishes (who first spell disallows the other and the only chance for Morgana is to act at the same time as Merlin does, although this case is not considered in the puzzle).
In order to represent the Merlin’s temporal precedence as a preemption factor for the Morgana’s action,
we may be tempted to modify the previous network and replace it with the network shown in Figure \ref{fig:trump}, right. But causal Bayesian networks are governed by the following postulate: if evidence is provided about a
node, the causal arrows reaching it are deactivated but all that start from it remain active. Instantiating the
node $S_{Me}$ means breaking the relation with the node Me but maintaining the causal flux from $S_{Me}$ to $S_{Mo}$
and $F_P$, even if the arrow is excitatory in both cases and it does not cancel the node $S_{Mo}$ as a possible
influence of $F_P$. In causal networks inhibitory links have no representation. ‘Acting before’ is key to correctly interpret that puzzle, but inhibitory arcs and temporal precedence has no representation in standard causal Bayesian networks. Petri Nets contribute to solve those deficiencies.

\section{A Fuzzy Stochastic Timed Petri Net approach.}

A Petri net (PN) is a graphical tool for modeling dynamic processes. A PN \cite{tadao1990petri} is a DAG showing the following components. Tokens representing resources in a broad sense, being physical or intangible, denoted by black dots. Places are locations where tokens are stored waiting to be transferred, denoted by a small circle. A circle containing a dot represents a place containing a token. Lastly, transitions, depicting changes in the status of the places and their tokens, representing actions and places, conditions. Places are connected with transitions by directed arrows. Let $t$ be a transition. Each place $p$ having an arrow from $p$ to $t$ is an input place of $t$. Each place $p$
having an arrow from $t$ to $p$ is an output place of $t$. Places are marked with tokens. A transition is enabled, for a given marking, if and only if all its input places have at least one token. Once a transition is fired, a token from each of its input places is removed and a token to each of its output places is added. So, if a transition is enabled, a new marking is reached, performing the dynamic behavior of the net.

Formally, a PN is a quintuple $PN=(P, T, I, O, M_0 )$, where:
$P=\{p_1 , p_2 , ..., p_m \}$ is a finite set of places.
$T=\{t_1 , t_2 ,...,t_n \}$ is a finite set of transitions, $P\cup T=\emptyset$.\\
$I: (PxT)\to \mathbb{N}$, is an input function defining directed arcs from places to transitions. $I(t_i , p_j )$ represents the number of arcs connecting a place $p_j$ with a transition $t_i$.
$O: (TxP)\to \mathbb{N}$, is an output function defining directed arcs from transitions to places. $O(t_i , p_j)$ represents
the number of arcs connecting a transition $t_i$ with a place $p_j$.
Parallel arcs connecting a place to a transition or vice versa are represented by a single directed arc
labeled with its weight, $w$. If $w=1$, the arc is not labeled.
$M_0: P\to \mathbb{N}$ is the initial marking.

Marking is the number of token in places. Once a PN is executed, the number and positions of places
dynamically change according to the transition firing, governed by the enabling rule and the firing rule,
both managing the flows of tokens in the net:

A transition $t$ is enabled if the number of tokens of each input place $p$ of $t$ is greater than or
equal to the weight of the directed arc connecting $p$ to $t$. On the other hand, a transition $t$ is enabled if the number of tokens of each input place $p$ of $t$ is greater than or equal to the weight of the directed arc connecting $p$ to $t$.

A transition without any input place is a source transition and one without any output place is a sink
transition. A source transition is unconditionally enabled and the firing of a sink transition consumes but does not generate any token.

Consider the following PN illustrated in Figure \ref{fig:petri_net}, left:

\begin{figure}[htbp]
\centering{
        \includegraphics[width=3cm]{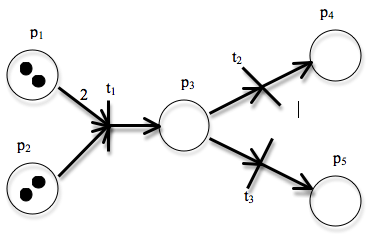}
        \includegraphics[width=3cm]{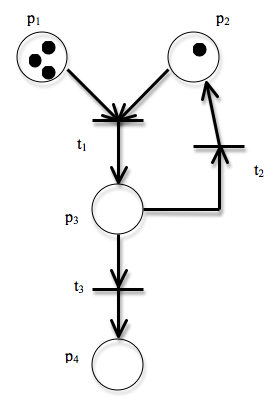}
}
\caption{Petri net representations.}
\label{fig:petri_net}
\end{figure}

The initial marking $M=(p_1 , p_2, p_3 , p 4 )$ of the PN corresponding to figure 12 is $M_0 =(2, 2, 0, 0)$. $t_1$ is the only
transition enabled. Firing it, we reach a new marking $M_1 =(0, 1, 1, 0, 0)$. Now, $t_2$ or $t_3$ can be fired. If $t_2$ is
activated, the new marking is $M_2 = (0, 1, 0, 1, 0)$ and if $t_3$ is fired, $M_3 =(0, 1, 0, 0, 1)$.

PN can be used to model realistic problems as the next one of the job offer and demand. Imagine a
situation in which there is a permanent job offer and three demands of employment, represented by Figure \ref{fig:petri_net}, right.

The initial marking is $M_0 =(3, 1, 0, 0)$ and $t_1$ is the only transition enabled. Firing it, a new marking $M_1$ is reached: $M_1 =(2, 0, 1, 0)$. Now, $t_2$ and $t_3$ are enabled, getting the new marking $M_2 =(2, 1, 0, 1)$. Enabling $t_3$ assures a permanent job offer to every possible demand as the loop suggest. It can be easily seen that the described Petri Nets can model prototypical scenarios of dynamic systems like the following ones: representing a sequential execution where transition $t_2$ can be enabled only if another transition $t_1$ is fired. The temporal constraint ‘$t_1$ precedes $t_2$ ’ or ‘$t_2$ after $t_1$ ’ can be modelled. They can model a case of conflict: both transitions are
enabled by the place $p_1$, but the firing of one of them disables the other. In this case, assigning probabilities is a usual way to decide in case of conflict. Petri Nets can also symbolize concurrency, \textit{e. g.}, processes that cooperate to achieve a
common goal: the firing of a transition $t_1$ can put a token on two places $p_2$ and $p_3$. Lastly, we can model synchronization: Let a transition $t_1$ be enabled only if places $p_1$ and $p_2$ have a token, hence modeling the joining operation.
Sowa firstly used basic PN in causal representation \cite{sowa1999knowledge}, modeling the ‘Yale shooting problem’,
originally proposed by \cite{hanks1987nonmonotonic} in the context of non-monotonic temporal reasoning (Cf.
Hans and McDermott, 1986) and adapted by Sowa to be represented with PN. The puzzle refers to a
dynamic scenario involving two relevant properties: being loaded (something concerning to a gun) and
being alive (something concerning to a victim) and two actions performed in sequence: wait and shoot.
The initial situation is that the gun is loaded and the victim is alive and a kind of law of inertia is
assumed: usually, properties of things do not change from an initial situation $s_0$ to others subsequent
situations $s_1$ , $s_2$ , etc. But the victim may die if in a subsequent situation the gun is fired. Nevertheless, non-monotonic logics do not lead to that conclusion because they disregard the relevance that the causal dependencies have in defeasible knowledge. Sowa aimed to use a single PN to show the relations
between the properties and actions involved in the Yale shooting problem:

\begin{figure}[htbp]
\centering{
        \includegraphics[width=6cm]{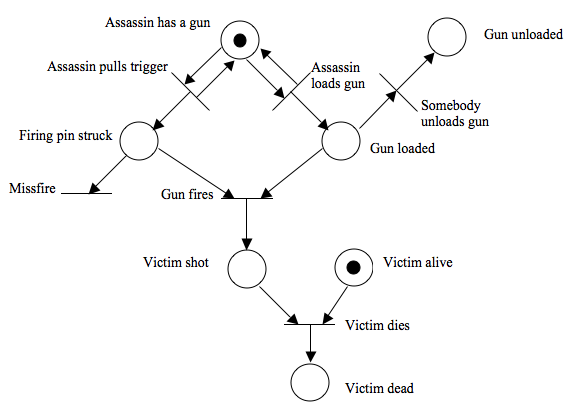}
}
\caption{Petri Net modeling the Yale shooting problem (redrawn from fig. 4.20 of Sowa \cite{sowa1999knowledge})}
\label{fig:gps}
\end{figure}

Next, we use PN to represent some of the aforementioned causal puzzles. As concurrence in Petri Net are attached to transitions, we will refer the simultaneous presence of causes with the ‘co-occurrence’ word.

\subsection{Co-occurrence and overdetermination.}

\begin{figure}[htbp]
\centering{
        \includegraphics[width=4cm]{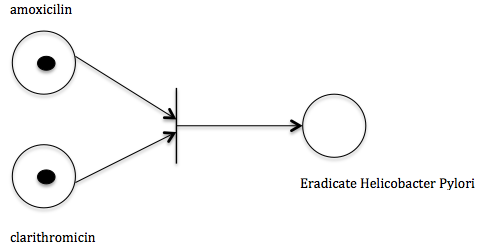}
        \includegraphics[width=4cm]{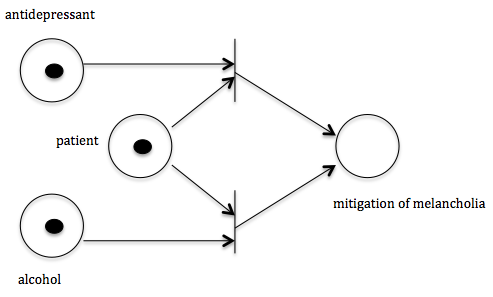}
}
\caption{Representing co-occurrence of places with PN (left). Representing asymmetrical over-determination as conflict in a Petri Net. (right)}
\label{fig:coas}
\end{figure}

Causal overdetermination can be illustrated using PN as a case of synchronization. Recall that in
symmetrical overdetermination two different causes must contribute to provoke the desired effect. For example, to eradicate the Helicobacter Pylory bacterium, clarithromycin and amoxicillin
should be jointly administered (Figure \ref{fig:coas}, left).

In order to the transition be fired, this PN requires the sincronization of both tokens, becoming
dependent places. Remember that in asymmetrical over-determination two causal factors contribute to provoke the effect, but they are incompatible each other and the first triggering the action inhibits the other. This puzzle can be represented as a case of conflict in Petri Nets (Figure \ref{fig:coas}, right).

Inhibition with Petri Nets is modeled inserting an extra node (labeled ‘patient’). To provide a remedy,
two ‘medicines’ and a patient should be considered. If a medicine is used by the patient, the patient node
loses its token and the other possible transition is deactivated. In this case, delays on transitions are not
required and the puzzle becomes adequately represented using a single PN. Note that, unlike Bayes Nets, dependence or independency in PN is a property of the transitions, not of
the nodes. In PN, independency is largely related with concurrence. A transition is independent from
other if it can be triggered before, after or at the same time; it is dependent if it depends on which other node or transition is enabled.

\subsection{Alternative causes}

Recalling the example of section 2.2, the voters contributing to the final decision were summarized by
the Boolean formula: $(A\wedge (B\lor C\lor D)) \lor (B\wedge C \wedge D)$, meaning that a surgery decision is performed if A votes
positively, and B or C or D votes positively or if B and C and D votes positively. The PN representing the
$(A\wedge B) \lor (A\wedge C) \lor (A\wedge D) \lor (B\wedge C\wedge D)$ denoting that several alternative clusters of necessary causal
factors are sufficient to provoke the action. Disjunctive causes have a no transparent representation in PN, because even if we can represent every cluster as a synchronization of causal factors, the surgery can be caused by the concurrency of all of them $(A\wedge B\wedge C\wedge D)$. So, the attempt to represent that case with a basic Petri Net is incomplete, because a dummy transition connecting the existing ones would be required. Basic PN considers neither time, nor probability, even if time and indetermination have a prominent role
in discrete-event systems involving causal links. Nevertheless, Merlin and Ramchandani \cite{bowden2000brief} separately extended ordinary PN for including time delay in two different ways:
firing and enabling durations.

In a basic PN, if a transition is enabled, it can be fired, moving input tokens from one place to
another. But if time is considered, the transition will have a time delay. This means that input
tokens are instantly transferred, but output tokens are not generated until the time delay is not
surpassed. When a transition fires, input tokens are instantly transferred and output tokens just generated,
but are not in disposition to enable new transitions until the delay associated to the target place is
exceeded.

In fact, firing and enabing durations are similar ways to represent time in a Petri Net. The difference is only about the delay
is positioned, whether in transitions or places. When time is assigned to a place, we denote the amount of
time that the tokens generated by the transitions are off for enabling new transitions. When time is
allocated in a transition, each input token has the same delay and is the transition delay what decides
when the output tokens emerged. In the sequel, we will consider time associated to transitions. Time gets
representation in Petri Nets extending the classic PN definition to the Timed Petri Nets (TPN):
$TPN=(P, T, I, O, M_0 , \tau)$, where P, T, I, O and $M_0$ is as in PN and $\tau$: $T \to R^+$ is a function that associates
transitions with time delays.
TPN enables the representation of time in a negative causal loop.

\subsection{Time delays in causal loops}

Recall the example of the three orders of sales that are made in a shop. That fact is represented in a PN putting three tokens in the input place (sales). So, a transition $t_1$ may be fired as many times as tokens are. Let us suppose that the delay associated to the
transition $t_1$ is null (\textit{e.g.}, when a sale is made, an order is executed) and the delay of transition $t_2$ is $4$, denoting $4$-days standby in applying the order. Then, as the orders are made effective three days after the
sales, the store may be out of supply, stopping the sales.

\subsection{Probability, time, trumping and fizzling.}

Regarding the representation of the trumping and fizzling puzzle, both of them involve time, co-
occurrence and also vagueness or indetermination.
Indetermination is introduced in Petri Nets using probabilities. Stochastic Timed Petri Net (STPN) is an
evolution of TPN dealing with probability in timed transitions \cite{molloy1985discrete} \cite{wang2007petri}. In a STPN, a
transition is qualified with a probable firing delay value. Formally, $STPN=(P, T, I, O, M_0 , \tau, \wedge)$, where P,
T, I, O, $M_0$ , $\tau$, is as in TPN and $\wedge$: $T\to R$ is a function that associates stochastic probability delays to
transitions.

\begin{figure}[htbp]
\centering{
        \includegraphics[width=4cm]{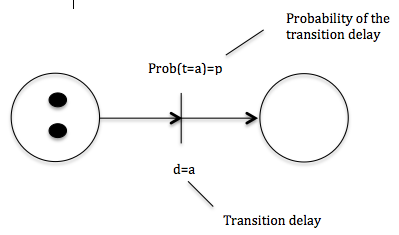}
        \includegraphics[width=4cm]{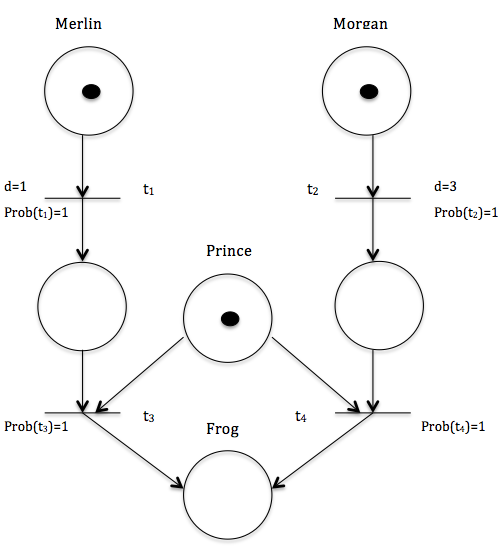}
}
\caption{Representation of time and imprecision in a STPN (left). Representing trumping as a conflict in a Petri Net with delays and probabilities. (right)}
\label{fig:time_trump}
\end{figure}

Next, we will show how to represent time and imprecision involved in the trumping and fizzling puzzle using a STPN showing a conflict case. Regarding trumping, it can be modeled with a STPN as a situation of conflict assigning probabilities and delays to transitions. The STPN shown in Figure \ref{fig:time_trump} is a case of conflict labeled with time delays and probability marks.

We remain to represent the fizzling. Fizzling is a causal puzzle concerning a pluri-causal
situation in which the effective cause is not the one that provides the greater probability to the effect: $a$ is
a quiet and responsible person and $b$ is a vandal. $a$ anticipates $b$ being the cause of the breakage of the
lamppost. The fizzled disposition of $b$ is not the cause of the effect, although his presence highly
increases the probability of it. ‘Highly’ is a vague probability qualification and fuzzy set theory provides
a way to measure that imprecise estimation.
Although both deal with imprecision, crisp and fuzzy probability differ. Meanwhile probability theory
deals with randomness, fuzzy set theory approaches vagueness. Fuzzy set theory probabilities can be numerical, based on fuzzy events or linguistic.
First, regarding numerical probability, Zadeh's formula \cite{zadeh1996fuzzy} makes probability equal to the integral, or a sum in the discrete case, of an expectation. Regarding linguistic probability, the unit interval values must be distributed; \textit{i.e.}, we have to fix a membership function to each probability predicate or by a function. Thus, fuzzy set theory provides tools to manage imprecise probabilities as ‘highly probable’ and fuzzy Petri Nets might benefit of its management, for example, using them as thresholds to trigger transitions in the fizzling puzzle.
The fizzling puzzle involves fuzzy probabilities and, then, a Fuzzy Stochastic Timed Petri Net (FSTPN)
is required to model it. A FSTPN assigns fuzzy probabilities to delays in
transitions. Formally, $FSTPN=(P, T, I, O, M_0 , \tau, \mu)$, where P, T, I, O, $M_0$ and $\tau$ is as in STPN and $\mu$:T
$\to R^+ \cup \Gamma$ is a function that associates to each transition a real value in $R^+$ or a linguistic label \cite{nguyen2018first}, suggesting that the underlying probability to the transition is approximate rather than
crisp.
The FSTPN shown in Figure \ref{fig:fstpn} illustrates how to represent time and fuzzy indetermination, characteristic of the fizzling puzzle:

\begin{figure}[htbp]
\centering{
        \includegraphics[width=4cm]{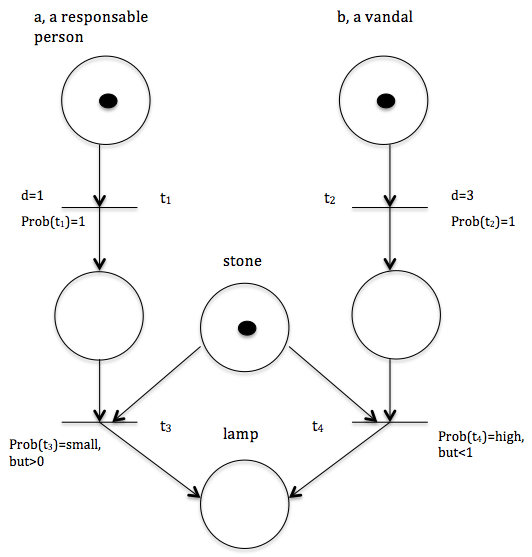}
}
\caption{Representing fizzling as a conflict in a Petri Net with delays and fuzzy probabilities}
\label{fig:fstpn}
\end{figure}

\section{Concluding remarks}

Although regular graph methods are useful for dealing with causal attributes, they fail to represent co-occurrence, time, circularity and fuzzy indetermination together. To overcome these difficulties, we pointed at the Fuzzy Stochastic Timed Petri Nets as a tool that can
contribute to represent all of them. Petri Nets and its extensions seems to be particularly appropriate for representing material causation,
based on the view that causes are physically connected to their effects and the causal link transfer a mark from the cause to the effect. The movements of tokens in PN from one place to another through transitions illustrate this flow from cause to effect. However, this interpretation presents some difficulties, some related to causality and others specific to the PN.

Regarding a causal scenario, negative causality \cite{schaffer2000causation} is a major objection. In some cases the absence of a fact causes the effect and so, nothing is transferred though the causal channel. In that case, no mark is transmitted from the cause to the effect; it is rather the absence that causes the effect. PN are not able to represent negative causality: only if there are tokens moving in the net, the dynamic of the event is reflected.
Pluricausality and multiple effects present also problems to be modeled in PN, because the representation of disjunctive events has redundancies. Even if it is possible to represent alternative causes in PN, the semantic of the net does not have a direct reading. And the same goes for multiple effects.

\bibliographystyle{acm}
\bibliography{main}

\end{document}